\documentclass[wcp]{jmlr}

\usepackage{wrapfig}

\usepackage{tikz}
\usepackage{pgfplots}

\usepackage{multirow}

\usepackage{longtable}

\usepackage{booktabs}
\usepackage[load-configurations=version-1]{siunitx} 

\title[Subspace Restricted Boltzmann Machine]{Subspace Restricted Boltzmann Machine}

  \author{\Name{Jakub M. Tomczak} \Email{jakub.tomczak@pwr.edu.pl}\\
   \Name{Adam Gonczarek} \Email{adam.gonczarek@pwr.edu.pl}\\
   \addr Institute of Computer Science\\Wroclaw University of Technology\\Wroc\l aw, Poland}

\begin{document}

\maketitle

\begin{abstract}
The subspace Restricted Boltzmann Machine (subspaceRBM) is a third-order Boltzmann machine where multiplicative interactions are between one visible and two hidden units. There are two kinds of hidden units, namely, \textit{gate units} and \textit{subspace units}. The subspace units reflect variations of a pattern in data and the gate unit is responsible for activating the subspace units. Additionally, the gate unit can be seen as a pooling feature. We evaluate the behavior of subspaceRBM through experiments with MNIST digit recognition task, measuring reconstruction error and classification error.
\end{abstract}

\section{Introduction}
\label{sec:intro}

The success of machine learning methods stems from appropriate data representation. Clearly this requires applying feature engineering, i.e., handcrafted proposition of a set of features potentially useful in the considered problem. However, it would be beneficial to propose an automatic features extraction to avoid any awkward preprocessing pipelines for hand-tuning of the data representation. Deep learning turns out to be a suitable fashion of automatic representation learning in many domains such as object recognition, speech recognition, natural language processing, or domain adaptation \citep{BCV:13}.

Fairly simple but still one of the most popular deep models for unsupervised feature learning is the Restricted Boltzmann Machine (RBM). The bipartie structure of the RBM enables block Gibbs sampling which allows formulating efficient learning algorithms such as contrastive divergence \citep{H:02}. However, lately it has been argued that the RBM fails to properly reflect statistical dependencies \citep{RKH:10}. One possible solution is to apply higher-order Boltzmann machine \citep{S:86} to model sophisticated patterns in data.

\begin{figure}[!htbp]
  \begin{center}
    \includegraphics[width=.3\textwidth]{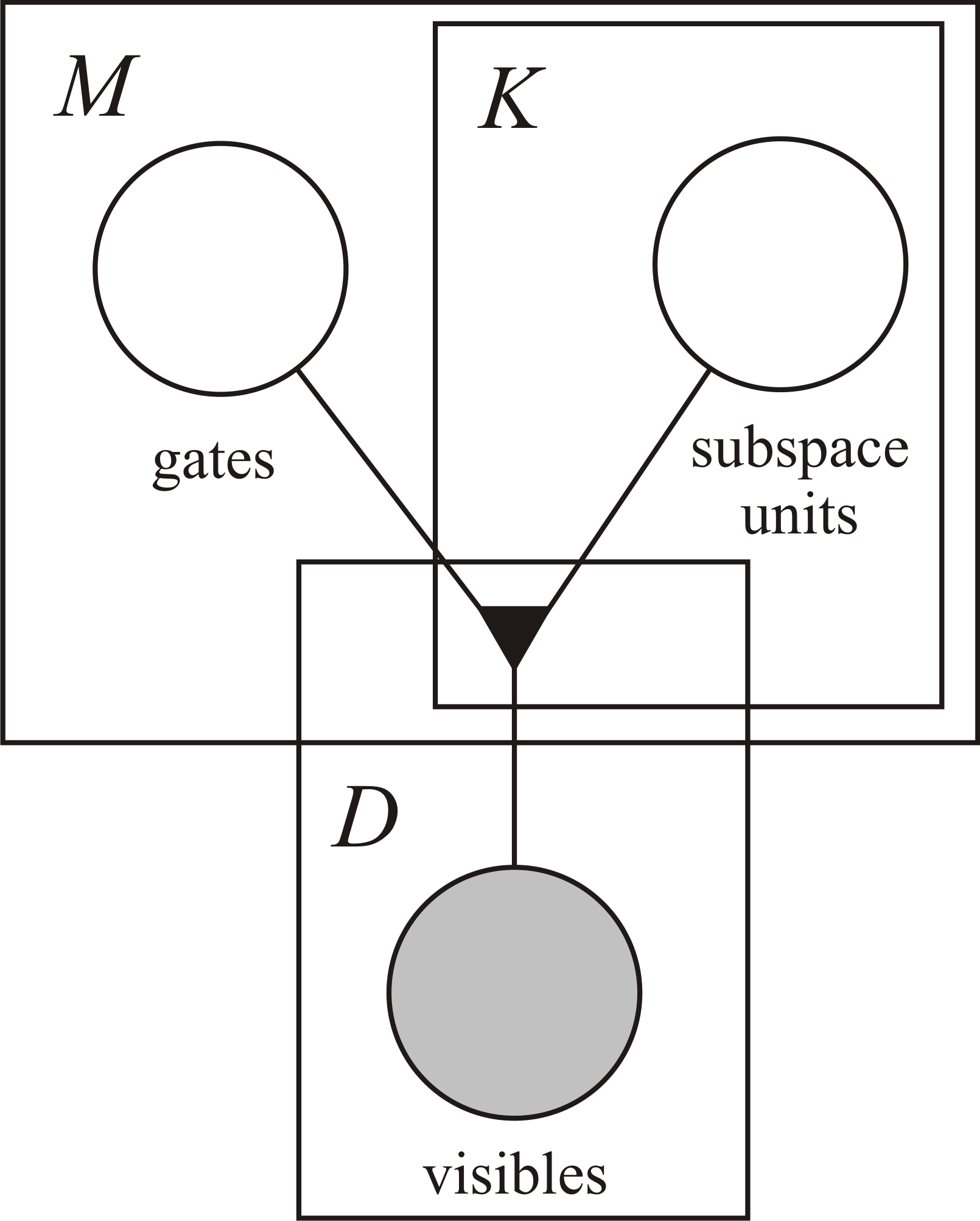}
  \end{center}
  \vspace{-10pt}
  \caption{A graphical representation of the subspaceRBM. The triangular symbol represents a third-order multiplicative interaction.}
  \label{fig:subspaceRBM}
\end{figure}

In this work we follow this line of thinking and develop a more refined model than the RBM to learn features from data. Our model introduces two kinds of hidden units, i.e., \textit{subspace units} and \textit{gate units} (see Figure \ref{fig:subspaceRBM}). The subspace units are hidden variables which reflect variations of a feature and thus they are more robust to invariances. The gate units are responsible for activating the subspace units and they can be seen as pooling features composed of the subspace features. The proposed model is based on an energy function with third-order interactions and maintains the conditional independence structure that can be readily used in simple and efficient learning.

\section{The model}
\label{sec:subspacerbm}

The RBM is a second-order Boltzmann machine with restriction on within-layer connections. This model can be extended in a straightforward way to third-order multiplicative interactions of one visible $x_{i}$ and two types of hidden binary units, a gate unit $h_{j}$ and a subspace unit $s_{jk}$. Each gate unit is associated with a group of subspace hidden units. The energy function of a joint configuration is then as follows:\footnote{$\mathbf{x} \in \{0,1\}^{D}, \mathbf{h} \in \{0,1\}^{M}, \mathbf{S} \in \{0,1\}^{M\times K}$}$^{,}$\footnote{The parameters are $\boldsymbol\theta = \{\mathbf{W}, \mathbf{b}, \mathbf{c}, \mathbf{D}\}$, where $\mathbf{W} \in \mathbb{R}^{D\times M\times K}$, $\mathbf{b}\in \mathbb{R}^{D}$, $\mathbf{c} \in \mathbb{R}^{M}$, and $\mathbf{D} \in \mathbb{R}^{M\times K}$.}
\begin{equation}\label{eq:energy}
E(\mathbf{x}, \mathbf{h}, \mathbf{S}|\boldsymbol\theta) = - \sum_{i=1}^{D} \sum_{j=1}^{M} \sum_{k=1}^{K} W_{ijk} x_{i} h_{j} s_{jk} - \sum_{i=1}^{D} b_{i} x_{i} - \sum_{j=1}^{M} c_{j} h_{j} - \sum_{j=1}^{M} h_{j} \sum_{k=1}^{K} D_{jk} s_{jk} .
\end{equation}
We refer the Gibbs distribution with the energy function in (\ref{eq:energy}) to as the \textit{subspace Restricted Boltzmann Machine} (subspaceRBM). For the subspaceRBM the following conditional dependencies hold true:\footnote{$\mathrm{sigm}(a) = \frac{1}{1+\mathrm{exp}(-a)}$}$^{,}$\footnote{$\mathrm{softplus}(a) = \log\big{(}1+\mathrm{exp}(a)\big{)}$}
\begin{align}
p(x_i = 1 | \mathbf{h}, \mathbf{S}) &= \mathrm{sigm} \big{(} \sum_{j} \sum_{k} W_{ijk} h_{j} s_{jk} + b_i \big{)} , \label{eq:probX} \\
p(s_{jk} = 1 | \mathbf{x}, h_j) &= \mathrm{sigm} \big{(} \sum_i W_{ijk} x_i h_j + h_j D_{jk} \big{)} \label{eq:probS} , \\
p(h_j = 1| \mathbf{x}) &= \mathrm{sigm}\Big{(} -K\mathrm{log} 2 + c_j + \sum_{k=1}^{K} \mathrm{softplus} \big{(} \sum_i W_{ijk} x_i + D_{jk} \big{)} \Big{)} , \label{eq:probH}
\end{align}
which can be straightforwardly used in formulating a contrastive divergence-like learning algorithm. Notice that in (\ref{eq:probH}) a term $-K\mathrm{log} 2$ imposes a natural penalty of the hidden unit activation which is linear to the number of subspace hidden variables. Therefore, the gate unit is inactive unless the sum of softplus of total input exceeds the penalty term and the bias term.

We can get some insight into the role of the subspace hidden variables by considering the conditional distribution over $\mathbf{x}$ with $\mathbf{S}$ marginalized out:
\begin{equation}\label{eq:xMarginalizedS}
p(\mathbf{x}|\mathbf{h}) \propto \mathrm{exp}(\sum_i b_i x_i + \sum_j c_j h_j) \prod_j \prod_k \Big{(} 1 +  \mathrm{exp}\big{(} h_j ( \sum_i W_{ijk} x_{i} + d_{jk} )  \big{)} \Big{)} .
\end{equation}
It turns out that marginalizing out $\mathbf{S}$ entails the visible variables being conditionally dependent. Hence the subspace hidden variables model the covariance of the visible variables.

\section{Learning}
\label{sec:learning}

In training, we take advantage of the equations (\ref{eq:probX}), (\ref{eq:probS}), and (\ref{eq:probH}) to formulate an efficient three-phase block-Gibbs sampling from the subspaceRBM. First, for given data, we sample gate units from $p(\mathbf{h}|\mathbf{x})$ with $\mathbf{S}$ marginalized out. Then, given both $\mathbf{x}$ and $\mathbf{h}$, we can sample subspace variables from $p(\mathbf{S}|\mathbf{x}, \mathbf{h})$. Eventually, the data can be sampled from 
$p(\mathbf{x}| \mathbf{h}, \mathbf{S})$.

We update the parameters of the subspaceRBM using contrastive divergence-like learning procedure. For this purpose we need to calculate the gradient of the log-likelihood function. The log-likelihood gradient takes the form of a difference between two expectations, namely, over the probability distribution with clamped data, and over the joint probability distribution of visible and hidden variables. Analogously to the standard RBM, both the expectations are approximated by samples drawn from the three-phase block-Gibbs sampling procedure.

\section{Related works}
\label{sec:related}

The standard RBM can reflect only the second-order multiplicative interactions. However, in many real-life situations, higher-order interactions must be included if we want our model to be effective enough. Moreover, often the second-order interactions themselves might represent little or no useful information. In the literature there were several propositions of how to extend the RBM to the higher-order Boltzmann machines. One such proposal is a third-order multiplicative interaction of two visible binary units $x_{i}$, $x_{i'}$ and one hidden binary unit $h_{j}$ \citep{H:10, RKH:10}, which can be used to learn a representation robust to spatial transformations \citep{MH:10}. Along this line of thinking, our model is the third-order Boltzmann machine but with different multiplicative interactions of one visible unit and two kinds of hidden units.

The proposed model is closely related to the \textit{subspace spike-and-slab RBM} (subspace-ssRBM) \citep{CDBB:13} where there are two kinds of hidden variables, namely, \textit{spike} is a binary variable and \textit{slab} is a real-valued variable. However, in our approach both the spike and slab variables are discrete. Additionally, in the subspaceRBM the hidden units $\mathbf{h}$ behave as gates to subspace variables rather than spikes as in ssRBM.

Similarly to our approach, gating units were proposed in the \textit{Point-wise Gated Boltzmann Machine} (PGBM) \citep{SZLL:13} where chosen units were responsible for switching on subsets of hidden units. The subspaceRBM is based on an analogous idea but it uses sigmoid units only whereas PGBM utilizes both sigmoid and softmax units.

Our model can be also related to RBM forests \citep{LBT:10}. The RBM forests assume each hidden unit to be encoded by a complete binary tree. In our approach each gate unit is encoded by subspace units. Therefore, the subspaceRBM can be seen as a RBM forest but with flatter hierarchy of hidden units and hence easier learning and inference.

Lastly, the subspaceRBM but with the softmax hidden units $\mathbf{h}$ turns to be the implicit mixture of RBMs (imRBM) \citep{NH:08}. However, in our model the gate units can be seen as pooling features while in the imRBM they determine only one subset of subspace features to be activated. The subspaceRBM brings an important benefit over the imRBM because it allows the subspaceRBM to relfect multiple factors in data.

\section{Experiment}
\label{sec:experiment}

We performed the experiment using MNIST image corpora\footnote{\url{http://yann.lecun.com/exdb/mnist/}} with different number of training images (10, 100, and 1000 per digit, i.e., $N\in \{100, 1000, 10000\}$). Additionally, the validation set of 10,000 and test set of 10,000 images were used. We compared the subspaceRBM with the RBM for the number of gate units equal $M = 500$ and different number of subspace units $K \in \{3, 5, 7\}$, measuring reconstruction error, classification error, and mean number of active gate units. For classification the logistic regression\footnote{The $\ell_2$ regularization was applied with the regularization coefficient equal $\lambda \in \{0, 0.01, 0.1\}$.} was fed up with the probabilities of gate units, $p(h_{j}=1|\mathbf{x})$, as inputs. The learning rate was set to $0.01$ and minibatch of size $10$ was used. The number of iterations over the training set was determined using early stopping according to the validation set reconstruction error, with a look ahead of 10 iterations.

\vspace{10mm}

\begin{figure}[!htbp]
  \begin{center}
    \includegraphics[width=.75\textwidth]{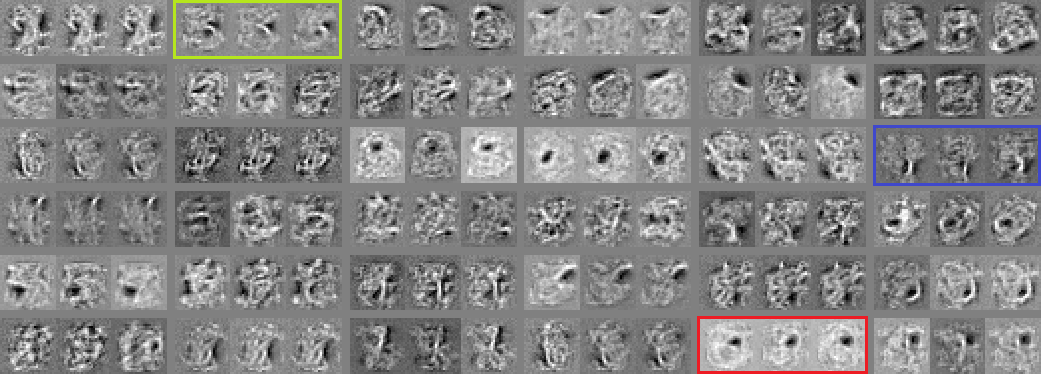}
  \end{center}
  \vspace{-10pt}
  \caption{Random subset of subspace features for $N=10000$ and the subspaceRBM with $M=500$ and $K=3$. Exemplary three groups of filters are outlined in red, blue and green which evidently tend to learn similar pattern with offsets in position, curvature or rotation.}
  \label{fig:subspace}
\end{figure}

\paragraph{Results and Discussion.} In Table \ref{tab:reconstruction} test reconstruction error is presented, and in Table \ref{tab:classification} -- test classification error. A random subset of subspace features is shown in Figure \ref{fig:subspace}. We notice that application of subspace units is beneficial for better reconstruction capabilities (see Table \ref{tab:reconstruction}). In the case of classification it is advantageous to use subspaceRBM in the case of small sample size regime (for $N$ equal 100 and 1000) with smaller number of subspace units. However, this result is rather not surprising because for over-complete representations simpler classifiers work better. On the other hand, for the small sample size there is a big threat of overfitting. Introducing subspace units to the hidden layer restricts the variability of the representation and thus preventing from learning noise in data. In the case of classification for larger number of observations, the best results were obtained for $K$ equal $5$ and $7$. This result suggests that indeed the subspace units lead to more robust features.

\begin{table}[!htbp]
\centering
\caption{Test reconstruction error for different settings of the RBM and the subspaceRBM evaluated on subsets of MNIST. The best results are in bold.}
\label{tab:reconstruction}
\begin{tabular}{c|c|c|c}
\hline
 & \multicolumn{3}{c}{\textbf{Reconstruction error}} \\
\cline{2-4}
\textbf{Model} & \textbf{\textit{N=100}} & \textbf{\textit{N=1000}} & \textbf{\textit{N=10000}} \\
\hline
RBM $M=500$ & 4.80 & 3.27 & 2.68 \\
subspaceRBM $M=500$, $K=3$ & 4.56 & \textbf{3.01} & 2.57 \\
subspaceRBM $M=500$, $K=5$ & \textbf{4.52} & 3.02 & 2.58 \\
subspaceRBM $M=500$, $K=7$ & 4.68 & 3.05 & \textbf{2.50}
\end{tabular}
\end{table}

\begin{table}[!htbp]
\centering
\caption{Test classification error for the RBM and different settings of the subspaceRBM evaluated on subsets of MNIST. The best results are in bold.}
\label{tab:classification}
\begin{tabular}{c|c|c|c}
\hline
 & \multicolumn{3}{c}{\textbf{Classification error [$\%$]}} \\
\cline{2-4}
\textbf{Model} & \textbf{\textit{N=100}} & \textbf{\textit{N=1000}} & \textbf{\textit{N=10000}} \\
\hline
RBM $M=500$ & 23.37 & 8.56 & 3.75 \\
subspaceRBM $M=500$, $K=3$ & \textbf{23.25} & 8.45 & 3.83 \\
subspaceRBM $M=500$, $K=5$ & 24.04 & \textbf{8.24} & 3.67 \\
subspaceRBM $M=500$, $K=7$ & 25.64 & 8.78 & \textbf{3.64}
\end{tabular}
\end{table}

\begin{table}[!htbp]
\centering
\caption{Number of active units for the RBM and different settings of the subspaceRBM evaluated on subsets of MNIST. The best results are in bold.}
\label{tab:classification}
\begin{tabular}{c|c|c|c}
\hline
 & \multicolumn{3}{c}{\textbf{Number of active units}} \\
\cline{2-4}
\textbf{Model} & \textbf{\textit{N=100}} & \textbf{\textit{N=1000}} & \textbf{\textit{N=10000}} \\
\hline
RBM $M=500$ & 78 & 68 & 54 \\
subspaceRBM $M=500$, $K=3$ & 82	& 112 & 49 \\
subspaceRBM $M=500$, $K=5$ & 74 & 88 & 78 \\
subspaceRBM $M=500$, $K=7$ & 46 & 72 & 64
\end{tabular}
\end{table}

\newpage
\section{Conclusion}
\label{sec:conclusion}

In this paper, we have proposed an extension of the RBM by introducing subspace hidden units. The formulated model can be seen as the third-order Boltzmann machine with third-order multiplicative interactions. We have showed that the subspaceRBM does not reduce to a vanilla version of the RBM, see equation (\ref{eq:xMarginalizedS}), and the subspace units incorporate a manner of modelling covariance of input variables. The carried-out experiments have revealed the supremacy of the proposed model over the RBM in terms of reconstruction error; in the case of classification error -- only for small sample size.

\bibliography{draft}

\end{document}